\documentclass[conference]{IEEEtran}
\IEEEoverridecommandlockouts
\newcommand{\todo}[1]{\color{red}TODO:\textbf{#1}\color{black}}
\usepackage{cite}
\usepackage{amsmath,amssymb,amsfonts}
\usepackage{graphicx}
\usepackage{textcomp}
\usepackage{xcolor}
\usepackage{comment} 
\usepackage{subcaption}
\usepackage{algorithm}
\usepackage{booktabs}
\usepackage{multirow}
\usepackage[utf8]{inputenc}
\usepackage{dblfloatfix} 
\usepackage[noend]{algpseudocode}

\makeatletter
\algnewcommand{\LineComment}[1]{\Statex \hskip\ALG@thistlm \(\triangleright\) #1}
\makeatother

\def\BibTeX{{\rm B\kern-.05em{\sc i\kern-.025em b}\kern-.08em
    T\kern-.1667em\lower.7ex\hbox{E}\kern-.125emX}}
    
\begin{document}

\title{Anomaly Detection in Video Data Based on Probabilistic Latent Space Models\\
%{\footnotesize \textsuperscript{*}Note: Sub-titles are not captured in Xplore and should not be used}
%\thanks{Identify applicable funding agency here. If none, delete this.}
}

\author{
  \IEEEauthorblockN{
    Giulia Slavic\IEEEauthorrefmark{1}, 
    Damian Campo\IEEEauthorrefmark{1}, 
    Mohamad Baydoun\IEEEauthorrefmark{1}, 
    Pablo Marin\IEEEauthorrefmark{2}, 
    David Martin\IEEEauthorrefmark{2},\\
    Lucio Marcenaro\IEEEauthorrefmark{1} and 
    Carlo Regazzoni\IEEEauthorrefmark{1}}
  \IEEEauthorblockA{
    \IEEEauthorrefmark{1}DITEN, University of Genova, Italy \\
      Contact email: slavic.giulia@gmail.com, damian.campo@edu.unige.it and mohamad.baydoun@edu.unige.it}
    \IEEEauthorblockA{\IEEEauthorrefmark{2}Intelligent systems lab, University Carlos III de Madrid, Spain\\
      Contact email: pamarinp@ing.uc3m.es and dmgomez@ing.uc3m.es}
    }
    
\iffalse
\author{
\IEEEauthorblockN{Giulia Slavic}
\IEEEauthorblockA{\textit{DITEN, University of Genova}\\
Genova, Italy \\
giulia.slavic@libero.it}
\and

\IEEEauthorblockN{Damian Campo}
\IEEEauthorblockA{\textit{DITEN, University of Genova}\\
Genova, Italy \\
damian.campo@edu.unige.it}

\and

\IEEEauthorblockN{Mohamad Baydoun}
\IEEEauthorblockA{\textit{DITEN, University of Genova}\\
Genova, Italy \\
mohamad.baydoun@edu.unige.it}

\and

\IEEEauthorblockN{David Martin}
\IEEEauthorblockA{\textit{Intelligent systems lab}\\
\textit{University Carlos III}\\
Madrid, Spain \\
mohamad.baydoun@edu.unige.it}
\and

\IEEEauthorblockN{Lucio Marcenaro}
\IEEEauthorblockA{\textit{dept. name of organization (of Aff.)} \\
City, Country \\
email address or ORCID}
\and

\IEEEauthorblockN{Carlo Regazzoni}
\IEEEauthorblockA{\textit{dept. name of organization (of Aff.)} \\
\textit{University of Genova}\\
Genova, Italy \\
email address or ORCID}

}
\fi
\maketitle

\begin{abstract}
This paper proposes a method for detecting anomalies in video data. A Variational Autoencoder (VAE) is used for reducing the dimensionality of video frames, generating latent space information that is comparable to low-dimensional sensory data (e.g., positioning, steering angle), making feasible the development of a consistent multi-modal architecture for autonomous vehicles. An Adapted Markov Jump Particle Filter defined by discrete and continuous inference levels is employed to predict the following frames and detecting anomalies in new video sequences. Our method is evaluated on different video scenarios where a semi-autonomous vehicle performs a set of tasks in a closed environment.
\end{abstract}


\begin{figure*}[t]
\begin{center}
\includegraphics[height = 4.5cm, width=\textwidth]{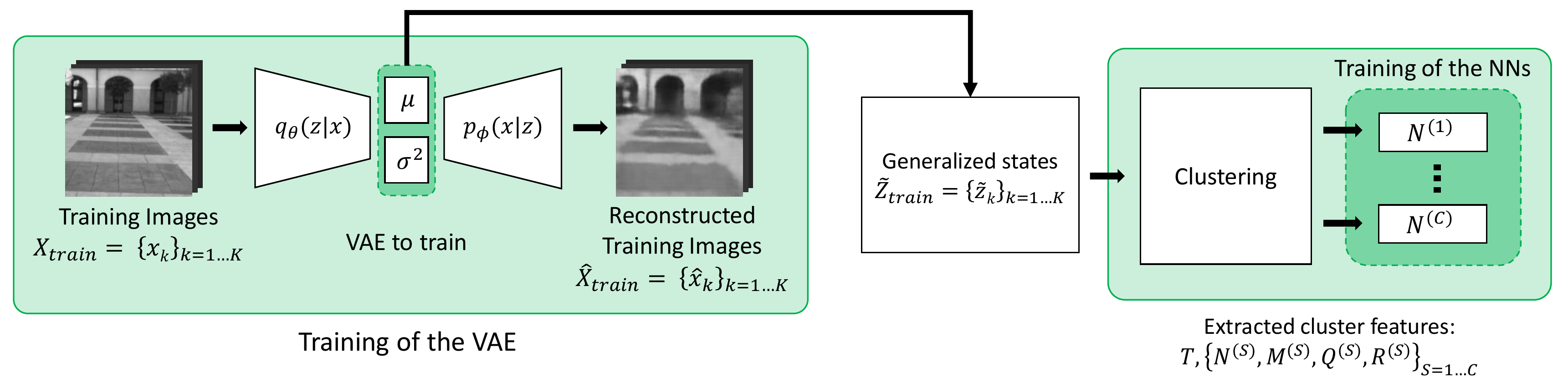}
%\vspace{-0.8cm}
\caption{Training phase: the VAE is trained to reconstruct a set of training images. The bottleneck features are then extracted and the GSs are derived and used for clustering. A Neural Network is trained for each cluster.}

\label{fig:training}
\end{center}
\end{figure*}

\section{Introduction}
The detection of anomalies on video data is currently one of the most relevant topics in signal processing and computer science fields due to its multiple applications such as machine automation \cite{Parisi2016,Rezazadegan2017,Lu2018,ko2015,Feng2018}, estimation of future instances \cite{Forkan2015,Bera2016,Ravanbakhsh2018a} and improvements in surveillance systems \cite{Cocsar2016,Bastani2016,Ravanbakhsh2019a}. Currently, video analysis and computer vision are important fields that attract large research and industrial interest. Furthermore, the automatic detection of anomalies in video information is a key element for generating robust autonomous systems that can adapt themselves to unknown situations/experiences and incrementally learn predictive models from them.

Over the last years, different deep learning algorithms have demonstrated their capabilities for solving several problems, e.g., image classification with a human-like performance \cite{Lake2013, Dodge2017,Geirhos2017}. Since we are moving closer and closer towards the ultimate aim of human-like vision for machines \cite{Nadeem2019}, more complex problems such as recognizing contextual information incrementally and understanding/adapt to new scenarios are current challenges to be automatically solved by machines. For generating autonomous systems, it is fundamental to provide machines with models that can handle the dynamic properties of real-world situations, e.g., dealing with streams of data that have not been seen before and including uncertainty in contextual representations and model predictions. 
%As pointed out in \cite{Rebuffi2017}, natural vision systems are inherently incremental: new visual information is continuously incorporated while existing knowledge is preserved. 

Motivated by the necessity of generating \textit{active} artificial agents that are able to interact with the real-world in often uncontrolled or detrimental conditions \cite{Sunderhauf2018}, this paper proposes a method for detecting anomalies in video data that facilitates the identification of unusual situations (previously unseen data) as they are experienced. Our method uses a probabilistic structure that facilitates the potential insertion/learning of new models as they are detected as abnormal, which would undoubtedly increase the adaptability of autonomous systems to unknown scenarios by continuously evolving their current models through the incorporation of new concepts. 

As is well known, in signal processing and surveillance systems, the detection of anomalies is an essential topic of large research \cite{Lin2015,Wang2018,Mabrouk2018,Sun2019,Ravanbakhsh2019a,Mahdyar2020} and commercial interest \cite{Aviv2011,Kiryati2014,Ishikawa2018}. Nonetheless, in computer vision, due to the large dimensionality of images, few attempts for recognizing and understanding anomalies have been made. Accordingly, existing methods based on deep learning for detecting anomalies in video sequences \cite{Ravanbakhsh2018a,Ravanbakhsh2018b, Wei2018,Li2019} usually do not allow to perform learning of models in a probabilistic framework that can be compatible with models relying on low dimensional information, e.g., 2-Dimensional position data. Novel research has tried to include the concept of \textit{lifelong learning} to deep neural networks (DNNs) \cite{Parisi2019}, which aim at allowing DNNs to acquire, fine-tune, and transfer knowledge through time continually. Nonetheless, advances in DNN lifelong learning remain in large dimensional representations that do not contemplate the possibility of including multisensory data for making robust models that could rely on either low or high dimensional information.

The proposed method is based on hierarchical probabilistic models that facilitate inferring future instances of video sequences and detecting anomalies. A similar approach was used for making inferences in low dimensional data, namely positional \cite{Baydoun_fusion} and control \cite{Kanapram2019} information coming from a moving vehicle. Accordingly, this paper proposes a method for inferring/estimating video sequences similarly as \cite{Baydoun_fusion} and \cite{Kanapram2019} do through a Markov Jump Particle Filter (MJPF). 

Our method is based on the latent representation of a Variational Autoencoder (VAE), which is employed for obtaining a low-dimensional state of the video at each time instant in a probabilistic fashion. Clusters of similar latent spaces are identified, which facilitates obtaining a semantic representation of video states. In each cluster, a fully connected neural network (NN) is employed to learn a non-linear dynamical model that allows estimating a future image given the content of the current one. Accordingly, two representation levels, i.e., discrete and continuous, are learned for making inferences in video sequences and detecting anomalies. For testing learned models, a particle filter coupled with a set of Unscented Kalman Filters (UKFs) \cite{Wan} are employed for predicting at discrete and continuous levels, respectively, and detecting anomalies. 

As mentioned before, the proposed method is based on research previously done on low-dimensional data \cite{Baydoun_fusion,Kanapram2019}. Nonetheless, the novel contributions of this paper are: \textit{i)} A full probabilistic method that represents video sequences into latent spaces so that video predictions can be made at continuous and discrete hierarchical levels. \textit{ii)} A method that models non-linearly the dynamics of latent space information through a set of NNs. \textit{iii)} Finally, an approach that is compatible with any low and large dimensional data, which allows us to detect anomalies in multisensory data and potentially use them for incremental learning.  

The rest of the paper is structured as follows: Section \ref{sec:method} presents the proposed method for detecting anomalies in video sequences. Section \ref{sec:dataset} introduces the dataset employed to evaluate our method. Section \ref{sec:results} shows and discusses obtained results. Finally, Section \ref{sec:conclusion} concludes the paper and presents some insights about possible future work. 

\section{Proposed method}\label{sec:method}
\begin{figure*}[t]
\begin{center}
\includegraphics[height = 4cm, width=\textwidth]{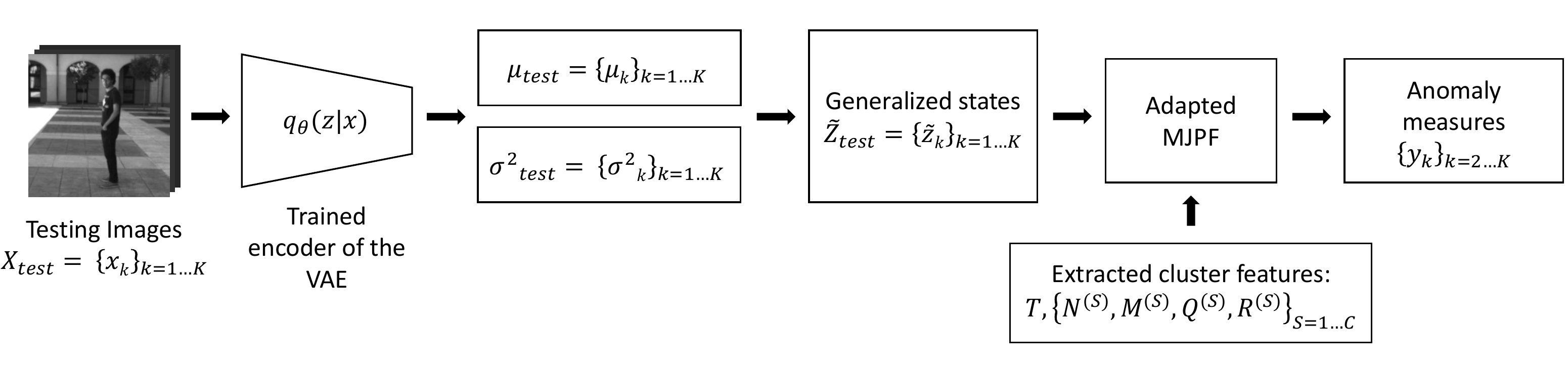}
%\vspace{-0.8cm}
\caption{Testing phase: the encoder of the VAE is used to extract the bottleneck features of the testing images. The GSs are derived and given as input to the Adapted MJPF, which detects anomalies by using information coming from clusters.}
\label{fig:testing}
\end{center}
\end{figure*}

The proposed method is composed of two basic steps, namely \textit{training} and \textit{testing phases}; during the former (section \ref{trainingPhase} and Fig. \ref{fig:training}), algorithms are trained based on observed data; whereas the latter (section \ref{testingPhase} and Fig. \ref{fig:testing}) uses learned algorithms to detect anomalies on new data. %Accordingly, detected abnormalities on the testing phase are employed to trigger new training phases that generate refined predictive models. Those new models are in turn added incrementally to a hierarchical probabilistic structure that is enriched as new situations are faced (section \ref{incremental}). 
%In the following sections we will first consider the training and testing of a single model alone, separated from the hierarchy, and then we will consider how the built models can be united together to form a hierarchy.

\subsection{Training phase}\label{trainingPhase}

\subsubsection{Variational Autoencoder}\label{VAE}

As a first step of the training phase, a VAE is used for describing the images in a latent space that has significantly reduced dimension with respect to the original image size. A VAE has been used instead of a normal Autoencoder because the former facilitates to represent images in the latent state probabilistically by using a mean $\mu$ and variance $\sigma^2$ to approximate each latent variable. This enables probabilistic reasoning and inference.

It is well known that a VAE is composed of two parts: an encoder $q_{\theta}(z|x)$ and a decoder $p_{\phi}(x|z)$. 
The latent state $z$ sampled from $\mathcal{N}( \mu,\sigma^{2} )$, returns an approximate reconstruction of the observation $x$. Through $\theta$ and $\phi$, we define the parameters of the encoder and decoder, respectively. In order to optimize them, the VAE maximizes the sum of the lower bound on the marginal likelihood of each observation $x$ of the dataset $D$, as described in \cite{kingma2014autoencoding,journals/ftml/KingmaW19}:

\begin{equation}\label{eq:opt1}
\mathcal{L}_{\phi, \theta}(D) = \sum_{x \in D}\mathcal{L}_{\phi, \theta}(x),
\end{equation}

being $\mathcal{L}(\theta, \phi; x)$ defined as:

\begin{equation}\label{eq:opt2}
\begin{aligned}
\mathcal{L}_{\phi, \theta,}(x) = & -D_{KL}({q_{\theta}(z|x) || p_{\phi}(z)}) + \\ 
& + E_{q_{\theta}(z|x)}[logp_{\phi}(x|z)], 
\end{aligned}
\end{equation}

where $D_{KL}$ defines the Kullback-Leibler divergence. Therefore, the first term measures the difference between the encoder's distribution $q_{\theta}(z|x)$ and the prior $p_{\phi}(z)$; being the prior typically a standard normal distribution $\mathcal{N}(0,1)$. The second term is the expected log-likelihood of the observation $x$ and forces the VAE to reconstruct the input data.

This work uses the ability of the VAE to encode the input information in a significant lower-dimensional space that exhibits probabilistic properties exploitable to detect anomalies at the latent feature level. Additionally, as we are interested in producing predictive models that can work for multisensory data regardless of their dimensionality, the VAE turns out to be an excellent choice for representing and treating video sequences as small-dimensional data, enabling a more homogeneous way of making algorithms for data fusion with multimodal information.

We first train the VAE by using a set of training images $X_{train}$. Then, we input again $X_{train}$ to the VAE and obtain a set of latent features described by $\mu_{train}$ and $\sigma^{2}_{train}$.

\subsubsection{Generalized states}\label{GSs}

Starting from the set of $\mu_{train}$, considered as the state of training images, we build a set of Generalized States (GSs) containing several time-order derivatives. This work only uses the first time-order derivative since no abrupt dynamics are considered.

Let $\mu_{k}$ be the value of $\mu$ for the image $x_k$ at time $k$, its first time-order derivative can be approximated by $\dot{\mu}_{k} \sim \frac{{\mu}_{k} - \mu_{k-1}}{\Delta k}$, where $\Delta k = 1$, which assumes a normalized regular sampling of images. The GS at time $k$ can thus be written as $\tilde{z}_{k} =[\mu_{k} \hspace{0.2cm} \dot{\mu}_{k}]^\intercal$. Repeating this for each consecutive couple of training images, we obtain a set of GSs for the training set, defined by:

\begin{equation}\label{eq:GSs_train}
\tilde{Z}_{train} = [\mu_{train} \hspace{0.2cm} \dot{\mu}_{train}]^\intercal.
\end{equation}

\subsubsection{Clustering and neural networks}\label{Clustering}

After obtaining GSs related to training video sequences, we use a traditional k-means algorithm to cluster GSs into groups that carry similar information. Since we use the values of $\mu$ and $\dot{\mu}$ to perform the clustering process, obtained clusters take into consideration the encoded image and also its dynamics w.r.t. the next frame. This facilitates recognizing and cluster different ways of moving, e.g., the vehicle crossing the same zone at different speeds.

Once the clustering is performed, a transition matrix $T$ encodes the transition probabilities from each cluster to the others. Additionally, the following features are extracted from each cluster indexed as $S$: \textit{i)} cluster's centroid $M^{(S)}$, \textit{ii)} cluster's  covariance $Q^{(S)}$ and \textit{iii)} cluster's radius of acceptance $R^{(S)}$. Finally, a fully connected neural network $N^{(S)}$ defining the dynamics of GSs, i.e., continuous predictive model, is learned for each cluster. Assuming that a total number of $C$ clusters have been identified, it is possible to write $S = \{1,\dots,C\}$. For training each $N^{(S)}$, the value of every $\mu_{k}$ is taken as input and the corresponding $\dot{\mu}_{k+1}$ as output, where $[\mu_k, \dot{\mu}_{k}]^\intercal \in S$. Moreover, to include the uncertainty of the Gaussian latent spaces encoded in $\sigma^2$, $2L$ additional inputs and outputs are used, where $L$ is the dimension of the latent state. Such $2L$ points, together with the initial mean $\mu^0_k = \mu_k$, permit to completely capture and define the Gaussian $\mathcal{N}( \mu_k,\sigma^{2}_k )$, as described in \cite{Wan}:

\begin{equation}\label{eq:sigma_points}
\begin{aligned}
\mu\mathrm{_k}^{i} & = \mu\mathrm{_k} + (\sqrt{(L+ \lambda)\Sigma\mathrm{_k}})\mathrm{_i}  \hspace{0.75cm} \textbf{if} \hspace{0.25cm} i = 1...L\\
\mu\mathrm{_k}^{i} & = \mu\mathrm{_k} - (\sqrt{(L+ \lambda)\Sigma\mathrm{_k}})\mathrm{_{i-L}} \hspace{0.4cm} \textbf{if} \hspace{0.25cm} i = L+1...2L,
\end{aligned}
\end{equation}

where $\lambda$ is a scaling parameter and $\Sigma_{k} \sim I_{L}\sigma^{2}_k$, being $I_L$ the identity matrix of dimension $L$. 

The $\mu_k^i$ values calculated in Eq. \eqref{eq:sigma_points} are the so-called sigma points associated with ($\mu_{k}$, $\sigma_k^2$). A corresponding group of sigma points can be defined in the same way for ($\mu_{k+1}$, $\sigma_{k+1}^2$) and each value of $\mu_{k+1}^i - \mu_{k}^i$ is given as additional output for the training of the NNs.

To summarize, each $N^{(S)}$ performs the following approximation:
$$ \dot{\mu}_{k+1}^i \sim N^{(S)}(\mu_{k}^i) + w_{k}^i,$$
where $\dot{\mu}_{k+1}^i$ and $\mu_{k}^i$ are calculated based on $[\mu_k, \dot{\mu_k}]^\intercal \in S$ and $w_{k}^i$ is the residual error after the convergence of the network.

Each $N^{(S)}$ learns a sort of \textit{quasi-semantic} information based on a particular image appearance and motion detected by the cluster $S$, facilitating the estimation of future latent spaces, i.e., predicting following frames. Such a feature can be employed to detect whether new observations are similar to previously learned situations encoded in the set of NNs. In case predictions from NNs are not compliant with observations, an anomaly should be detected, and models should be adapted to learn new situations, generating new semantic information.   

\subsection{Testing Phase}\label{testingPhase}
During the testing phase, each testing image is processed by the VAE, and GSs are calculated. Then, an adapted version of the MJPF is used to detect anomalies in video sequences.

\subsubsection{Adapted Markov Jump Particle Filter}\label{MJPF}
A MJPF can be described as a Probabilistic Switching Graphical Model \cite{Koller2009,Sucar2015} for prediction and anomaly detection purposes by using a bank of Kalman Filters (KFs) at the continuous state level and a Particle Filter at the discrete state level \cite{Baydoun_fusion}. This work tackles a problem that requires a non-linear model for prediction purposes and a non-linear observation model, solved by a set of NNs (each of them associated with a detected cluster) and a VAE, respectively. Accordingly, a bank of standard KFs at the state level cannot be used due to the nonlinearities described above. This work uses a bank of modified KFs whose predictions follow the same logic of the UKFs and employ the encoded information of the VAE for updating purposes.

\begin{figure}[t]
  \centering
  \centerline{\includegraphics[width = \linewidth]{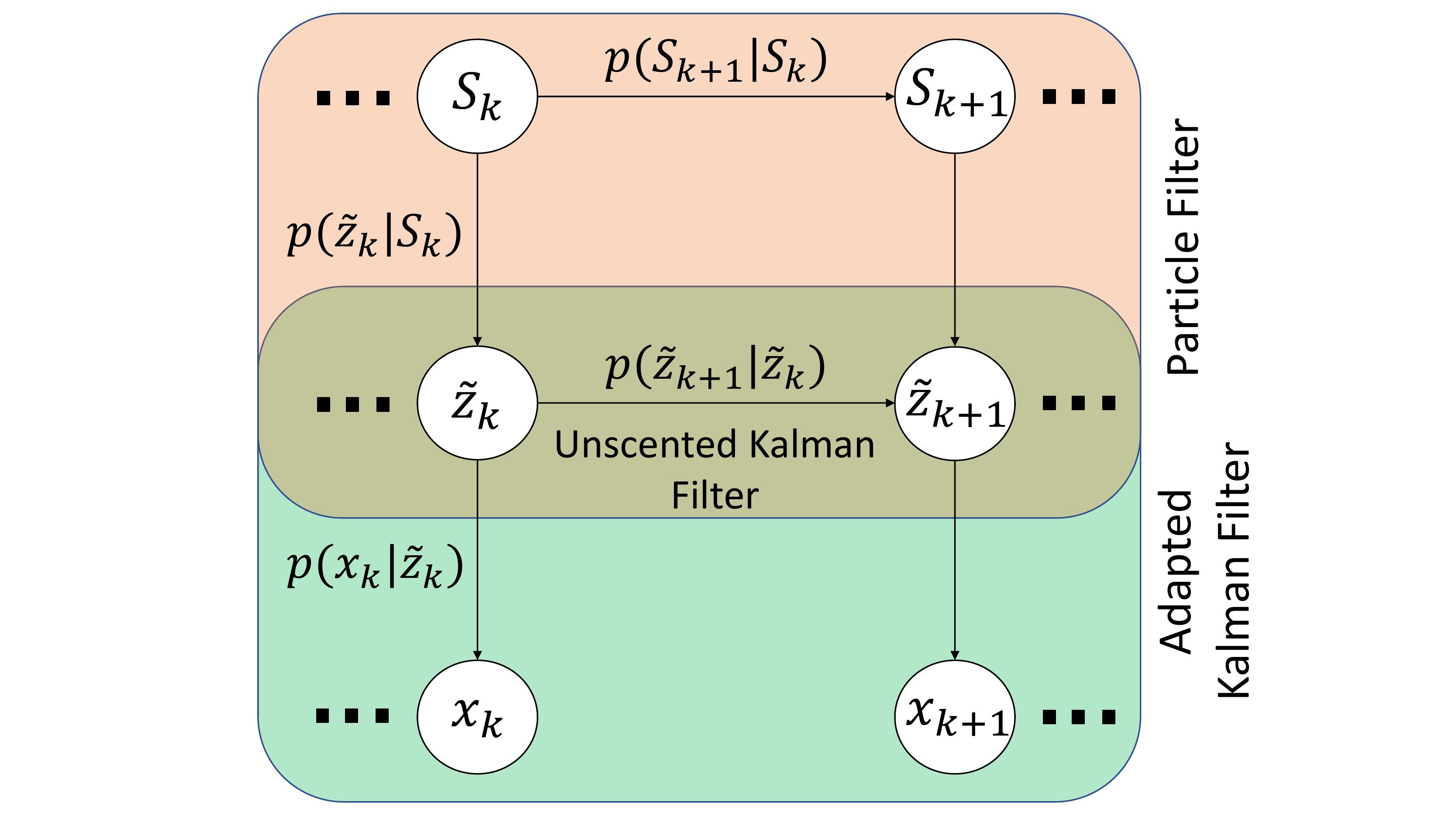}}
\caption{DBN associated with the A-MJPF.}
\label{fig:DBN}
\end{figure}

The DBN associated with the Adapted MJPF (A-MJPF) is showed in Fig. \ref{fig:DBN}. A detailed description of the MJPF can be found in \cite{Baydoun_fusion}. This paper will only provide a brief description of the overall logic of the employed probabilistic architecture, which is identical to the one used in \cite{Baydoun_fusion}. Instead, it will focus on the parts that have been modified in the A-MJPF w.r.t. the standard MJPF.

For both architectures (MPJF and A-MPJF), at each time instant $k$, two stages are performed: \textit{prediction} and \textit{update}. During prediction, the next cluster $S_{k+1}$ (discrete level) and the next GS $\tilde{z}_{k+1}$ (continuous level) are estimated for each particle, i.e., $p(S_{k+1} | S_{k})$ and $p(\tilde{z}_{k+1}| \tilde{z}_{k})$, respectively. Similarly to the standard MPJF, predictions at the discrete level in the A-MJPF are performed by using the transition matrix $T$ in each particle. On the other hand, predictions at the continuous level in the A-MPJF are performed by the neural network $N^{(S_k)}$ associated with the selected discrete state $S_k$. Since non-linear models are considered for predicting continuous level information, an UKF performs the estimations as described in \cite{Wan} by taking $2L$ additional sigma points as already done in section \ref{Clustering}. Therefore, the prediction for each sigma point follows the equation below:

\begin{equation}\label{eq:prediction}
\tilde{z}^{i}\mathrm{_{k+1}} = f(\tilde{z}^{i}\mathrm{_{k}}) = A\tilde{z}^{i}\mathrm{_{k}} + BN\mathrm{^{(S)}}(\mu^{i}\mathrm{_{k}}) + w^{i}\mathrm{_k},
\end{equation}

where $A$ and $B$ are two matrices used to map the previous state $\tilde{z}^i_k$ and the new velocity computed by $N^{(S)}(\mu^{i}_{k})$ on the new state $\tilde{z}^i_{k+1}$, such that as $A = [A_1 A_2]$ with $A_1 = [I_L 0_{L,L}]^\intercal$, $A_2 = 0_{2L, L}$ and as $B = [I_L I_L]^\intercal$.
The mean and covariance of the predicted state are then calculated using the UKF formulas for the propagation of a Gaussian random variable through a non-linear model.

The update phase is performed when a new measurement (image) is observed. At the discrete level, particles are resampled based on a measure of the anomaly (see section \ref{Abnormality}). At the state level, a modified version of the KF update is performed. This update takes into consideration the fact that $\mu_k$ and $\sigma_k^2$ from the VAE for image $x_k$ can be used as the mapped observation on the state space at time $k$. Consistently, $\sigma_k^2$ can approximate the covariance matrix, such that $\Sigma_{k} \sim I_{L}\sigma^{2}_k$, representing the uncertainty while encoding images. 
%\textcolor{red}{Supposing a negligible observation noise, it is possible to employ a modified version of the KF update equations where the observation matrix disappears(is not clear for me!)}. 
Algorithm \ref{algorithm_KF} describes the employed KF's steps.

\begin{algorithm} 
\caption{Equations of the prediction and update phase of the Adapted Kalman Filter.}
\begin{algorithmic}[1]
\LineComment \textbf{PREDICTION}: 
\State Calculation of the sigma points $\tilde{z}^i_{k|k}$ and of their respective weights $\tilde{W}^{i, m}$ and $\tilde{W}^{i, c}$ as described in \cite{Wan}. 
\State $\tilde{z}^i_{k+1|k} = f(\tilde{z}^i_{k|k})$ 
\State $\tilde{z}_{k+1|k} = \sum_{i = 0}^{2L}\tilde{W}^{i, m}\tilde{z}^i_{k+1|k}$ 
\State $P_{k+1|k} = \sum_{i=0}^{2L}\tilde{W}^{i,c}\{\tilde{z}^i_{k+1|k} - \tilde{z}_{k+1|k}\}\{\tilde{z}^i_{k+1|k} - \tilde{z}_{k+1|k}\}^\intercal$
\State $P^L_{k+1|k} = P_{k+1|k}\Big|_{\{row: 1...L, col: 1...L\}}$
\vspace{0.25cm}
\LineComment \textbf{UPDATE}: 
\State $K_{k+1} = [P^L_{k+1|k}; I_{L}] (P^L_{k+1|k} + \Sigma_{k+1})^{-1}$ 
\State $\tilde{z}_{k+1|k+1} = \tilde{z}_{k+1|k} + K_{k+1}(\mu_{k+1} - \mu_{k+1|k})$ 
\State$P_{k+1|k+1} = P_{k+1|k} - K_{k+1}(P^L_{k+1|k} + \Sigma_{k+1})K^\intercal_{k+1}$
\end{algorithmic}
\label{algorithm_KF}
\end{algorithm}

\subsubsection{Anomaly measurement}\label{Abnormality}
After the update phase, at each time instant $k$, the predicted value of $\mu_{k}^{l, p}$ related to latent state component $l$ and particle $p$ is compared with the actual updated value, outputting a measure of innovation defined as:
\begin{equation}\label{eq:abnormalityValue} 
y_k = \min_{p}\frac{\sum_{l = 1}^{L}\big|\mu_{k|k}^{l, p} - \mu_{k|k-1}^{l,p}\big|}{L}.
\end{equation}

The anomaly values of training video sequences are used to set an anomaly threshold defined as:
\begin{equation}\label{eq:threshold}
thresh = \bar{y}_{train} + 3std(y_{train}),
\end{equation}
being $\bar{y}_{train}$ and $std(y_{train})$ the mean value and standard deviation of anomalies from the training data, respectively. When using algorithm \ref{algorithm_KF} on testing data, video sequences that produce anomaly signals above the threshold in Eq.\eqref{eq:threshold} are considered as potential anomalies. Moreover, to filter out spurious peaks, also a temporal criterion based on a window of 3 frames is used, such that anomaly signals that are above the threshold $thresh$ but last less than 3 frames are not considered as actual anomalies.

\section{Employed Dataset}\label{sec:dataset}
\begin{figure}[H]
	\centering
	\begin{subfigure}[t]{0.22\textwidth}
	\includegraphics[width=3.6cm,height=2.5cm]{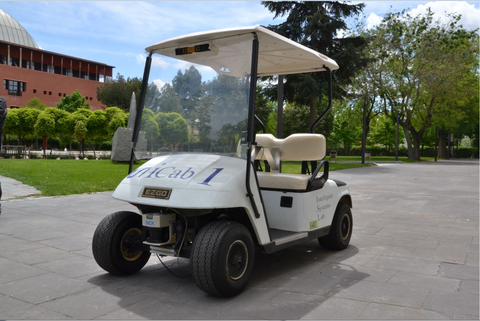}
	\caption[iCab]{Autonomous vehicle ``iCab''}
	\label{fig:iCab}
	\end{subfigure}
	~
	\centering
	\begin{subfigure}[t]{0.22\textwidth}
	\includegraphics[width=3.6cm,height=2.5cm]{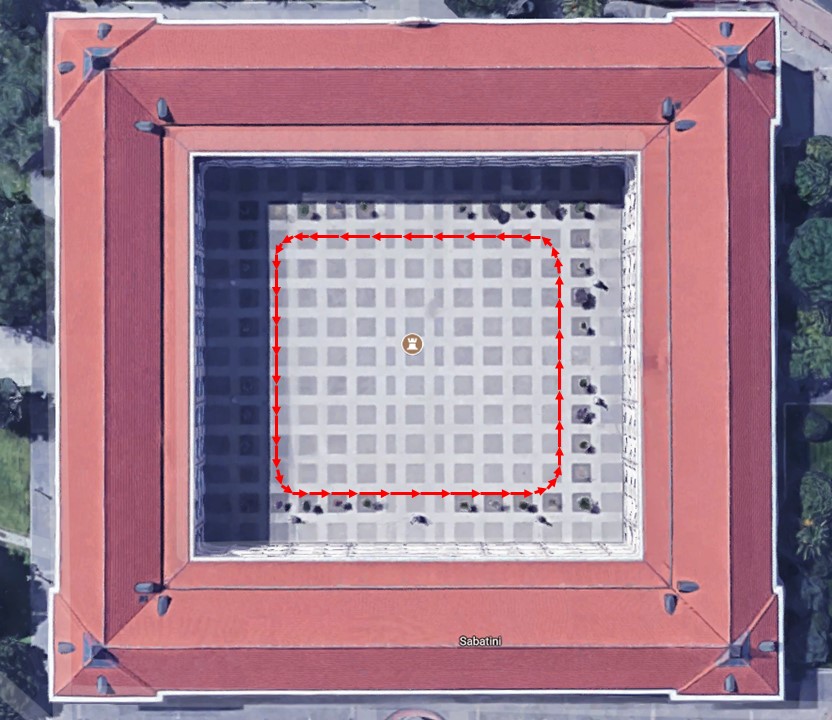}
	\caption[environment]{Closed environment}
	\label{fig:Environment}
	\end{subfigure}
\end{figure}

\begin{figure}[h]		
	\begin{subfigure}[t]{4cm}
		\centering
		\includegraphics[width=3.7cm,height=3.2cm]{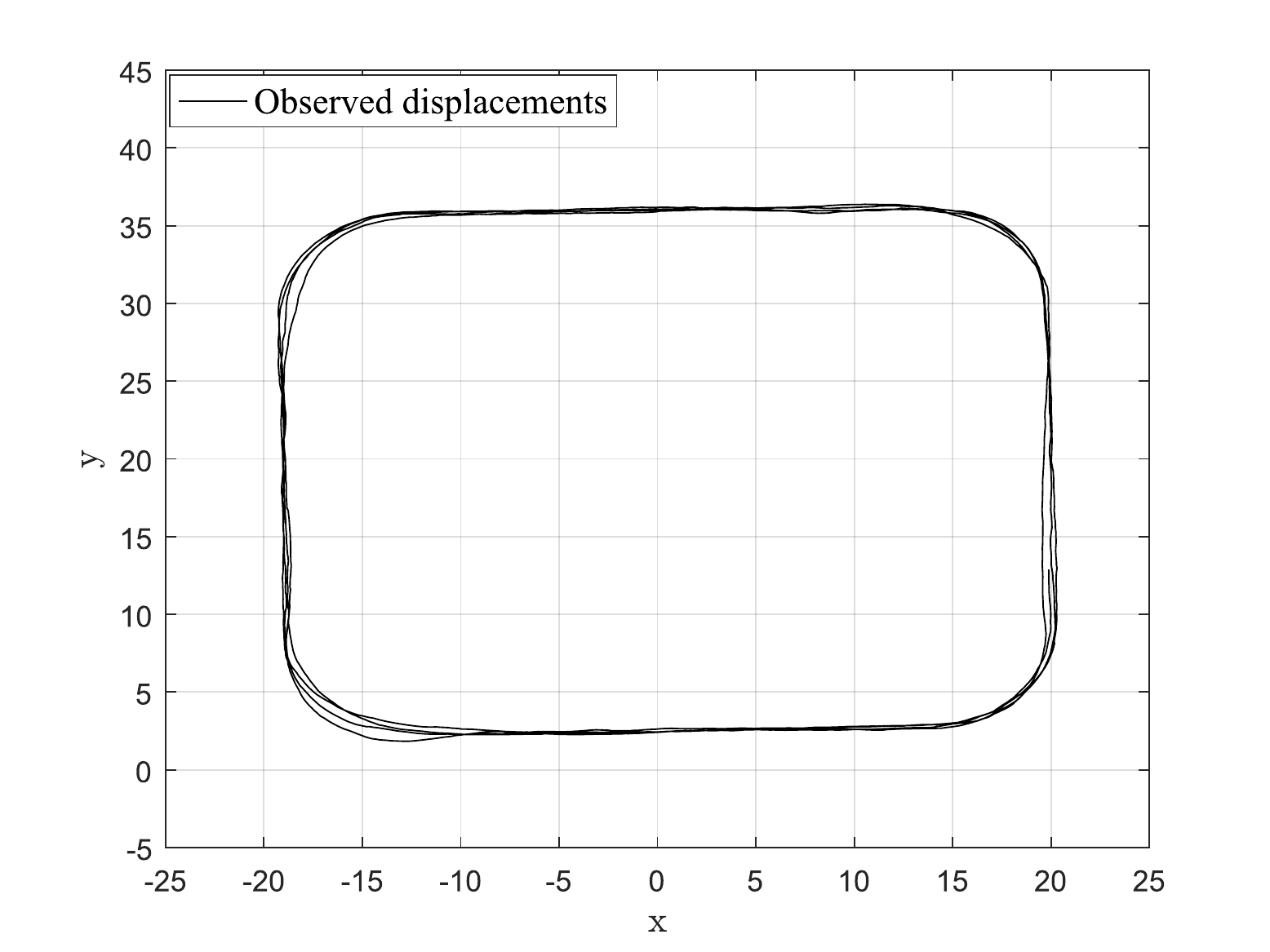}
	\caption{Perimeter monitoring}
		\label{fig:Monitoring}
	\end{subfigure}
	\begin{subfigure}[t]{4cm}
		\centering
		\includegraphics[width=3.7cm,height=3.2cm]{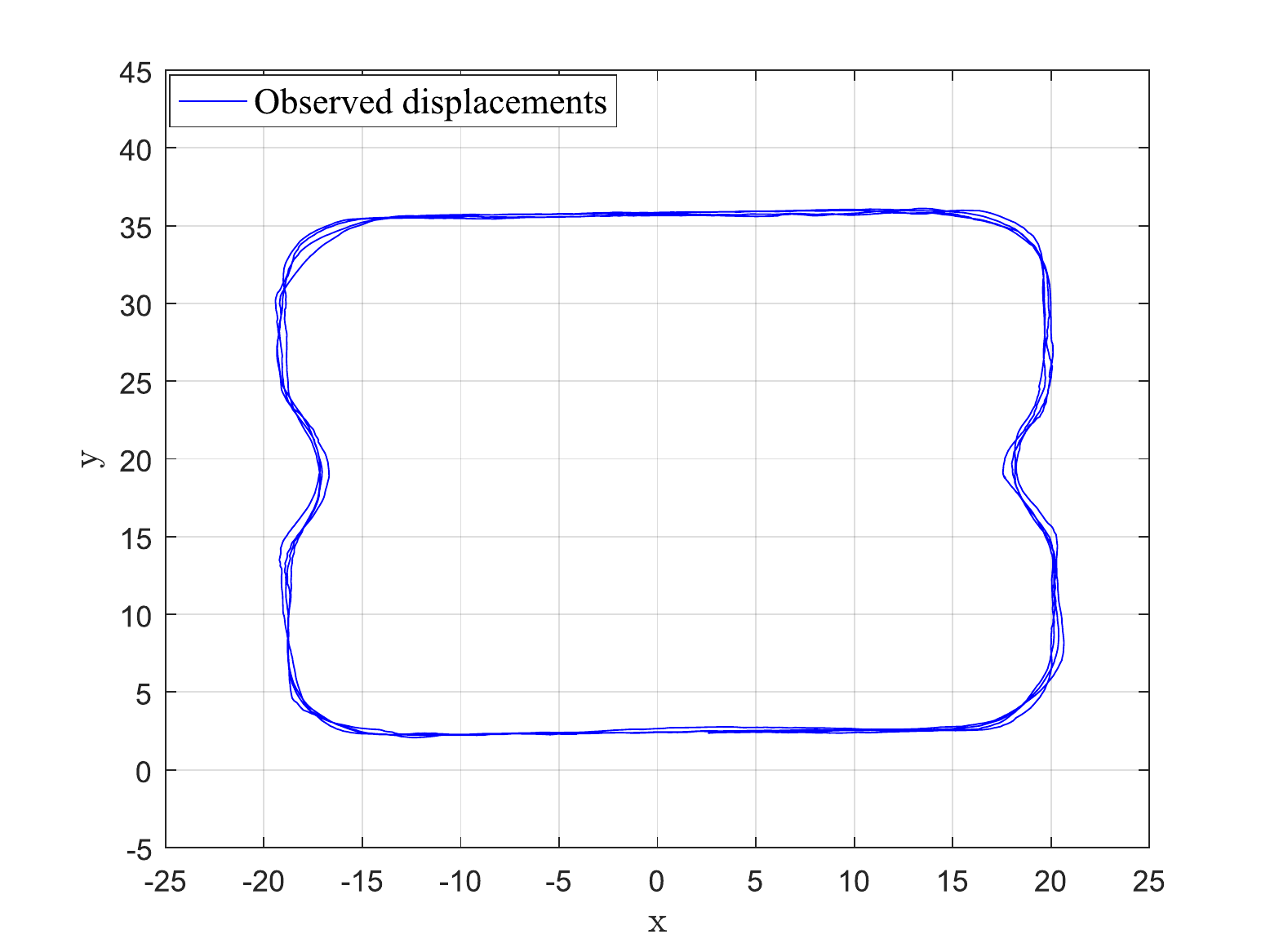}
	\caption{Pedestrian avoidance}
    \label{fig:Avoid_Dataset}
	\end{subfigure}
	\begin{subfigure}[t]{4cm}
		\centering
		\includegraphics[width=3.7cm,height=3.2cm]{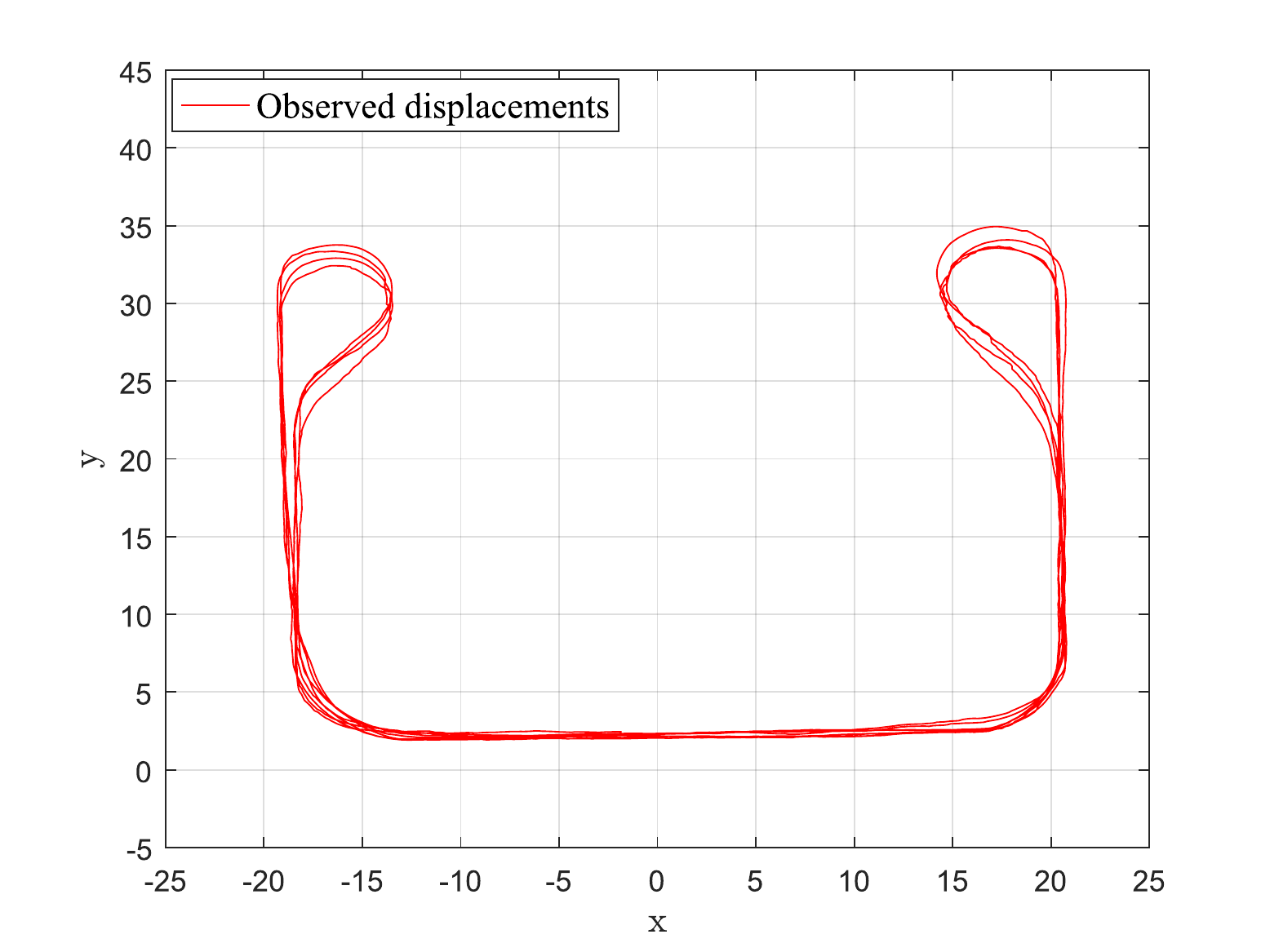}
	\caption{U-turn}
    \label{fig:U-Turn_Dataset}
	\end{subfigure}
	\hspace{0.75cm}
	\begin{subfigure}[t]{4cm}
		\centering
		\includegraphics[width=3.7cm,height=3.2cm]{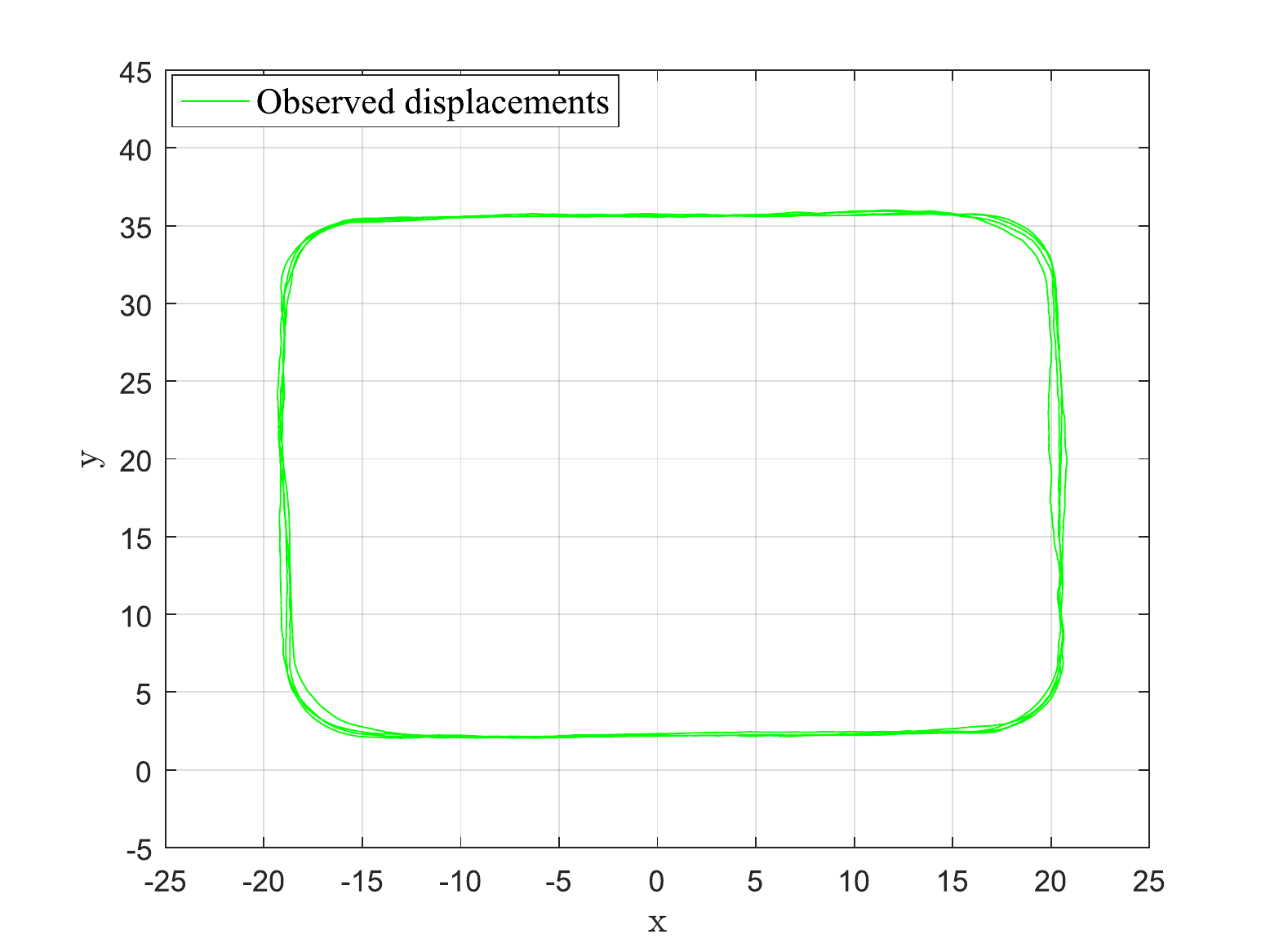}
	\caption{Emergency stop}
		\label{fig:Stop}
	\end{subfigure}
	\caption{Vehicle tasks used to evaluate the proposed method. Perimeter monitoring is utilized in the training phase whereas the other three tasks are employed for testing purposes}
	\label{fig:scenarios}
	\end{figure}

A real vehicle called ``iCab'' \cite{Marin2016}, see Fig. \ref{fig:iCab}, is used to collect the video dataset. A human drives the iCab performing different tasks in a closed environment displayed in Fig. \ref{fig:Environment}. The proposed dataset was captured from an onboard front camera while the vehicle executes 4 different tasks. 
We aim at detecting dynamics that have not been seen previously in a normal situation (Scenario I), which is used for learning purposes. Scenarios II, III and  IV include unseen maneuvers caused by the presence of pedestrians while the vehicle performs a perimeter monitoring task. Accordingly, four scenarios (see Fig.~\ref{fig:scenarios}) are considered in this work:\\

\begin{figure}[h]
  \centering
  \centerline{\includegraphics[width = \linewidth]{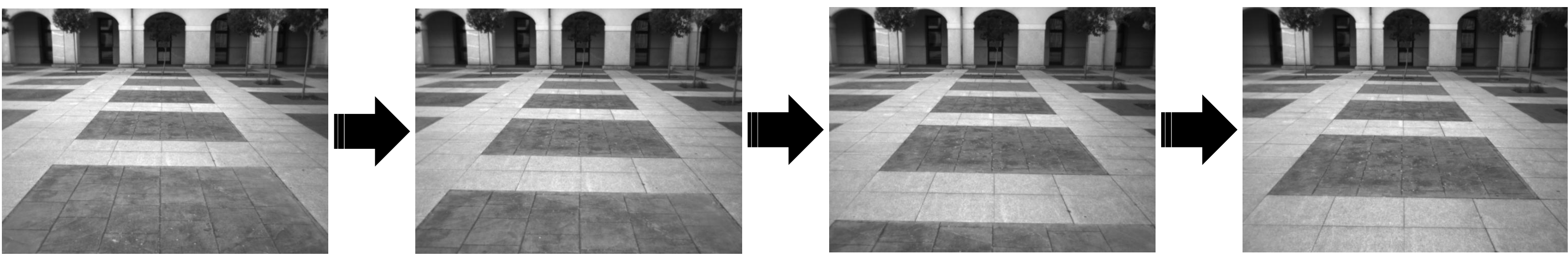}}
\caption{Scenario I (perimeter monitoring).}
\label{fig:pm}
\end{figure}

{\bf Scenario I (perimeter monitoring)}: the vehicle follows a rectangular trajectory along with a closed building. The temporal evolution of the perimeter monitoring maneuver from a first-person perspective is shown in Fig.~\ref{fig:pm}.
	
\begin{figure}[h]
  \centering
  \centerline{\includegraphics[width = \linewidth]{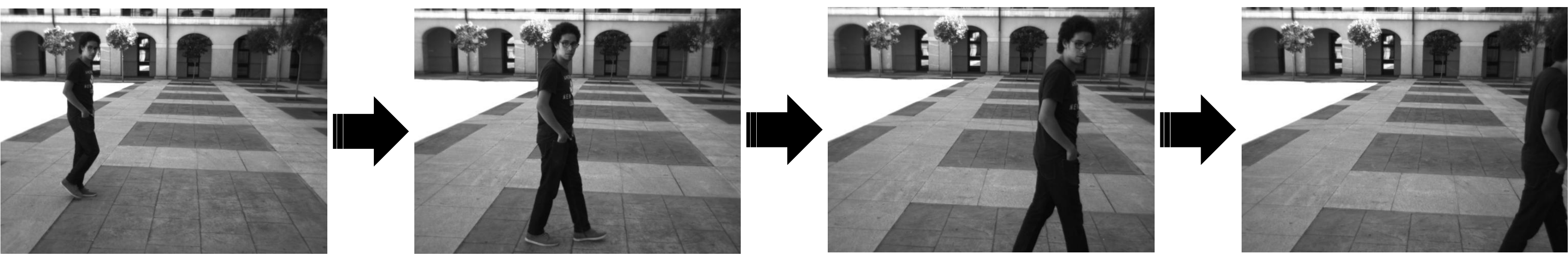}}
\caption{Scenario II (stop maneuver).}
\label{fig:stop}
\end{figure}

{\bf Scenario II (emergency stop maneuver)}: the vehicle executes the perimeter monitoring task and encounters two pedestrians crossing its path at each lap. The vehicle performs an emergency stop and then continues the perimeter monitoring task when pedestrians exit its field of view. A vehicle’s first-person perspective of the temporal evolution of the stop maneuver is provided in Fig.~\ref{fig:stop}.

\begin{figure}[H]
  \centering
  \centerline{\includegraphics[width = \linewidth]{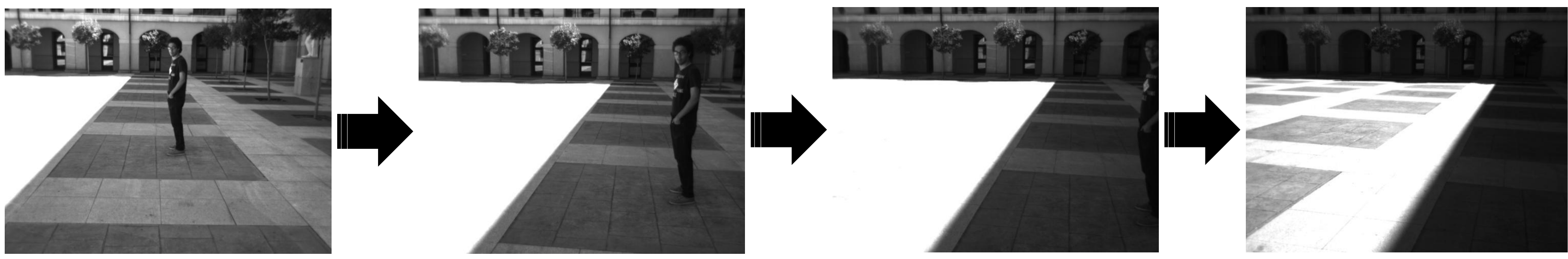}}
\caption{Scenario III (avoidance maneuver).}
\label{fig:avoid}
\end{figure}

{\bf Scenario III (pedestrian avoidance maneuver)}: two obstacles (stationary pedestrians) in different locations interfere with the perimeter monitoring task of Scenario I. The vehicle performs an avoidance maneuver and continues the perimeter monitoring. Fig.~\ref{fig:avoid} shows the temporal evolution of the avoidance maneuver from a first-person perspective.

\begin{figure}[H]
  \centering
  \centerline{\includegraphics[width = \linewidth]{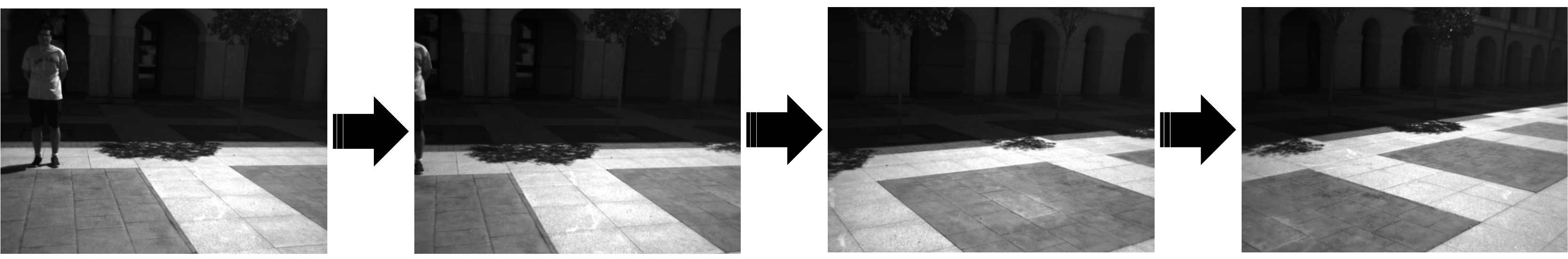}}
\caption{Scenario IV (U-turn maneuver).}
\label{fig:Uturn}
\end{figure}

{\bf Scenario IV (U-turn maneuver)}: while the vehicle executes a perimeter monitoring, it encounters two static pedestrians located in different locations. In this scenario, the vehicle performs a U-turn motion and then continues the perimeter monitoring in the opposite direction w.r.t. training data. Fig.~\ref{fig:Uturn} shows the U-turn maneuver from the front camera viewpoint.

\section{Experimental Results}\label{sec:results}

\begin{figure*}[t]
\centering
\begin{subfigure}[b]{\textwidth}
\rotatebox{90}{\hspace{1.3cm}(a)}
   \includegraphics[width=0.97\linewidth]{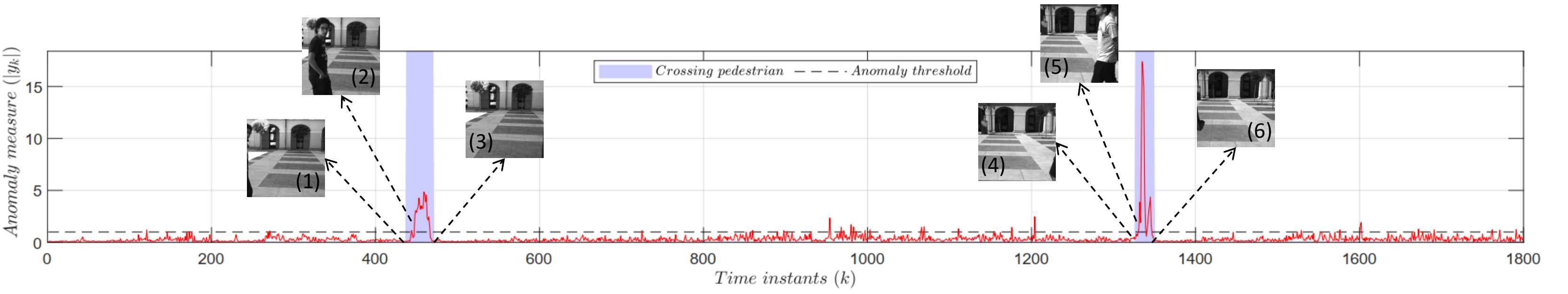}
\end{subfigure}
\hspace*{-3mm}
\begin{subfigure}[b]{1\textwidth}
\hspace{0.05cm}
\rotatebox{90}{\hspace{0.5cm}(b)}
   \includegraphics[width=0.97\linewidth]{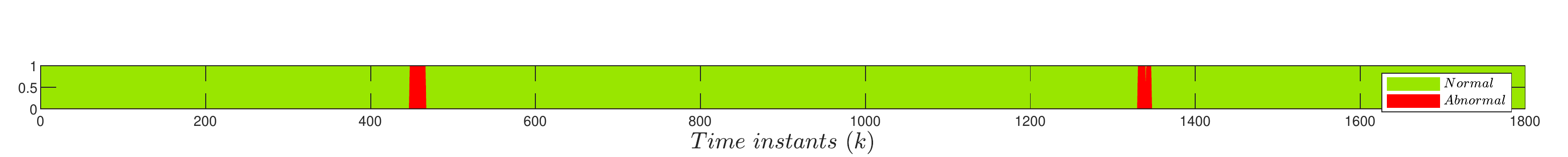}
\end{subfigure}

\caption{Testing phase on the emergency stop task: (a) Anomaly signal. (b) Color-coded final anomaly.}
\label{fig:abn_Stop}
\end{figure*}

\begin{figure*}[b]
\centering
\begin{subfigure}[b]{\textwidth}
\rotatebox{90}{\hspace{1cm}(a)}
   \includegraphics[width=0.97\linewidth]{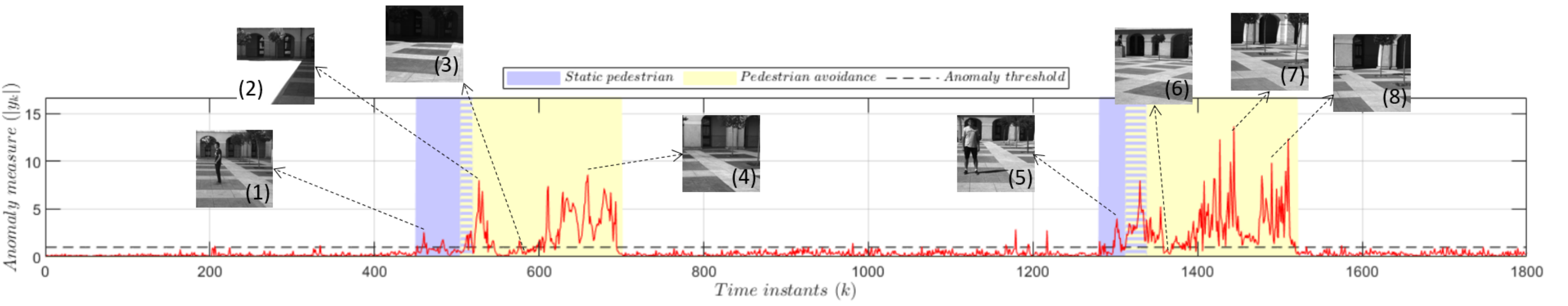}
\end{subfigure}
\hspace*{-3mm}
\begin{subfigure}[!b]{\textwidth}
\hspace{0.05cm}
\rotatebox{90}{\hspace{0.5cm}(b)}
   \includegraphics[width=0.97\linewidth]{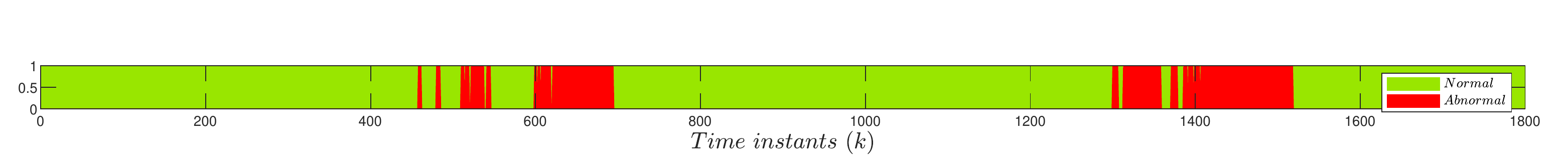}
   \label{fig:anomalyTimeThreshAMl}
\end{subfigure}

\caption{Testing phase on the pedestrian avoidance task: (a) Anomaly signal. (b) Color-coded final anomaly.}
\label{fig:abn_OA}
\end{figure*}

\begin{figure*}[t]
\centering
\begin{subfigure}[b]{\textwidth}
\rotatebox{90}{\hspace{2cm}(a)}
   \includegraphics[width=0.97\linewidth]{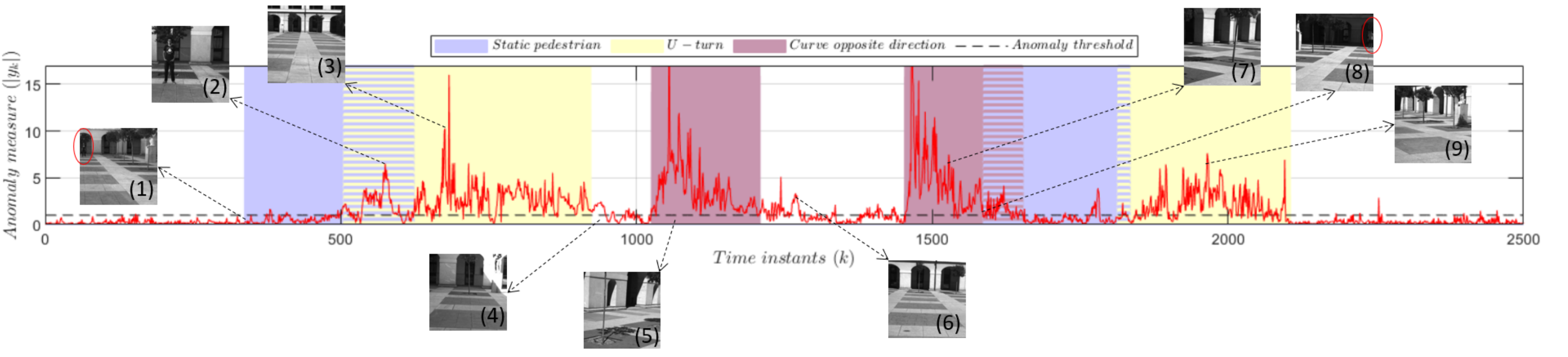}
\end{subfigure}
\hspace*{-3mm}
\begin{subfigure}[b]{\textwidth}
\hspace{0.05cm}
\rotatebox{90}{\hspace{0.5cm}(b)}
   \includegraphics[width=0.97\linewidth]{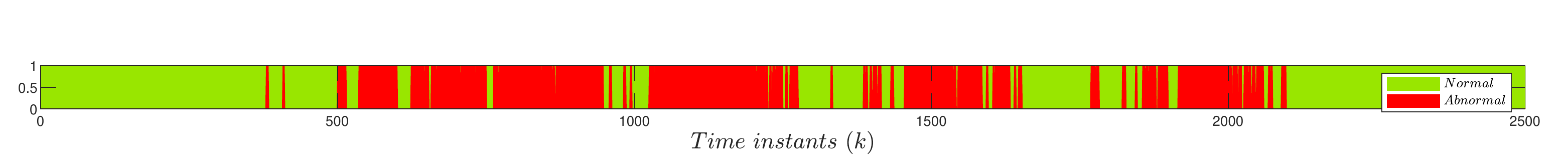}
\end{subfigure}

\caption{Testing phase on pedestrian avoidance through U-turn task: (a) Anomaly signal. (b) Color-coded final anomaly.}
\label{fig:abn_U_turn}
\end{figure*}

\subsection{Definition of normality}

As a first step, it is necessary to train our model to learn normal patterns. The Perimeter Monitoring frames are therefore used as $X_{train}$ data to perform the training phase described in section \ref{trainingPhase}. Consequently, a VAE is trained (\ref{VAE}), from which a set of clusters and their corresponding NNs (\ref{Clustering}) are learned based on frames where the iCab moves in a \textit{pedestrian-less} environment. We considered several cluster experiments where the total number of clusters $C$ was varied. Consequently, we selected the case of $C = 6$ empirically as it describes the normal scene accurately with a relatively low number of clusters. Video frames are clustered based on images' appearance and dynamics (changes in consecutive frames) at each time instant.

The threshold described in section \ref{Abnormality} is then obtained by performing a testing procedure using the training data. This threshold will be used in the actual testing phases described in the following sections with the objective of detecting abnormal behaviors in new scenarios.

\subsection{Emergency stop maneuver}
In this modality, the vehicle (artificial agent) performs an emergency stop maneuver that allows a pedestrian to cross in front of it. Fig. \ref{fig:abn_Stop}a displays the anomaly signal obtained from a single vehicle's lap around the courtyard where it encounters/interacts with two pedestrians. The signal is normalized based on the threshold calculated from the perimeter monitoring experience; see Eq.\eqref{eq:threshold}. Consistently, when the anomaly signal goes above the threshold (displayed as a dotted black line in Fig. \ref{fig:abn_Stop}(a)), a potential abnormal situation is detected. The blue color background indicates the presence of the pedestrian, such as from the moment it enters into the camera's field of view, see (1) and (4); until the instant where it leaves it, see (3) and (6). The anomaly signal grows rapidly as soon as the pedestrian completely enters the field of view of the camera, see (2) and (5).

Fig. \ref{fig:abn_Stop}(b) displays the final color-coded anomalies w.r.t. the perimeter monitoring task (training data), normal and abnormal frames are colored in green and red, respectively. It can be seen how the proposed method enables the detection of anomalies due to moving pedestrians that have not seen before in the training data. 

\subsection{Pedestrian avoidance}\label{avoidance}

In this modality, the vehicle avoids a static pedestrian. Fig. \ref{fig:abn_OA}a shows the resulting anomaly signal. The blue zones refer to video frames that contain the static pedestrian, and yellow zones encode the avoidance maneuvers. As can be seen from Fig. \ref{fig:Avoid_Dataset}, at each lap, the vehicle encounters two different static pedestrians in the environment. They wear t-shirts of different colors (black in the first case and white in the second one), which make them ``camouflage'' with the environment in some particular configurations due to changeable illumination conditions. This factor influences the different anomaly values for both pedestrians, with the second one generating a higher anomaly.

At each pedestrian encounter, the anomaly signal is composed of three zones with high values: a first one due to the pedestrian presence beginning, see (1) and (5); and other two due to the avoidance maneuver, see (2)-(4) and (6)-(8). Between the latter two peaks, a zone with a low anomaly is present, see (3), or (6); this is due to the execution of similar behaviors already observed in the training set.

\begin{comment}

\begin{figure*}[ht]
\centering

\includegraphics[width=1\linewidth]{figures/ABN_AM_3.pdf}
\caption{\todo{Is there any caption for this Figure? Please include (a),(b), (c), (d) at the left side with times roman font}}
\label{fig:ABN_AM_3} 

\end{figure*}

\end{comment}

\subsection{Pedestrian avoidance through U-turn}

In this modality, the vehicle avoids a static pedestrian by performing a U-turn maneuver. In this case, the vehicle will move in the opposite direction w.r.t. its previous motion towards the pedestrian, see Fig. \ref{fig:U-Turn_Dataset}, which introduces later on new situations, e.g., curving in the opposite direction, and other already known behaviors, e.g., moving straight in regions that present similar symmetries to those in the training set. Additionally, the U-turn presents some cases containing both normal and abnormal information, e.g., moving straight in regions that are similar to video sequences already seen but showing some differences related to structures that are illuminated differently in the courtyard.

Fig. \ref{fig:abn_U_turn}a shows the anomaly signal for this case. Three regions have been highlighted describing the main abnormal situations appearing in this task: \textit{i)} the presence of pedestrians colored in blue, \textit{ii)} the U-turn maneuver represented in yellow and \textit{iii)} the curves performed in the opposite direction w.r.t. the training set, which is coded in purple. Regarding the first situation, it must be noted that the blue zone is again related to the entire sequence of frames in which the pedestrian is present. In this task, as the pedestrian first appears on the opposite corner of the courtyard w.r.t. where the vehicle is moving, it occupies only a small amount of pixels, which makes its recognition difficult even for the human eye; see (1) and (8) in Fig. \ref{fig:abn_U_turn}a. Moreover, as pedestrians in this task are located on a background that has a similar color shade to their clothing, even when the vehicle is moving closer, low anomaly levels are detected. In (2), the anomaly is particularly high because the vehicle starts moving right to perform the U-turn.  The other two anomaly zones regarding the U-turn maneuver, see (3) and (9), and the curves in the opposite direction (5) and (7) are both detected effectively by the proposed method as anomalies. 

In addition to the three anomaly cases mentioned above, other smaller anomaly peaks can be observed in zones where the vehicle moves straight but in the opposite direction to what is experienced in the training set. Those anomalies due to different causes: In (6), they are caused by differences in the background (e.g., the right side is abnormal due to the presence of a tree and the lack of shadows). In (4), anomalies are found due to differences in the background and velocities (e.g., the vehicle is in the middle of the courtyard, where it is expected the maximum velocity based on the training dataset, but due to the U-turn maneuver, the vehicle decelerates at that section of the courtyard).

\subsection{Results Discussion}

It can be observed how the proposed method is able to determine whether a situation is normal or abnormal, with a high accuracy level in different scenarios. The detection of anomalies was particularly good in three cases: \textit{i)} visual data that differs substantially from training samples, e.g., when performing a curve in the opposite direction w.r.t. the training video sequences or during the central part of the U-turn movement. \textit{ii)} new image motions related to visual information that has already been seen, e.g., the beginning of the U-turn maneuver (due to the changing of direction) and after exiting from it (due to previously unseen velocity changes). \textit{iii)} the presence of moving or static objects, e.g., pedestrians, that are not considerably far away from the camera. On the other hand, the accuracy is much lower in cases where anomalies are not completely defined, e.g., pedestrians placed far away from the camera or that ``camouflage'' with the background. 

We have therefore observed how our method successfully recognizes anomalies based on the appearance (new images) and the dynamics (abnormal motions) of video data.

%It can be observed how through our method we were able to determine if a situation was normal or abnormal with a good level of accuracy that however varied depending on the situation itself: very good accuracy in case of motion changes or abnormal images due to a motion change and much lower accuracy in the case of a compressed local abnormality like the presence of a pedestrian. This second situation however had in itself some limitations due to the conformation of the pedestrian image (especially in the case of the U-turn). After obtaining the abnormalities we created a new model that was able to learn the new situations. Combined with the old one, it was possible to discriminate if a situation belonged to the first set of knowledge (Perimeter Monitoring) or to the second set of knowlegde (Emergency Stop, Pedestrian avoidance or U-turn), creating therefore a high-level semantic that combines itself with the low-level semantic given by the clusterization over the different scenarios.

\section{Conclusion and Future work} \label{sec:conclusion}

This paper presented a method for the detection of anomalies from visual data. By using a VAE, we were able to bring high-dimensional data acquired from a camera to a low-dimensionality that is compatible with other sensor data acquired from the vehicle and already examined in previous contributions (e.g., position, steering angle). An A-MJPF has been introduced and used to detect anomalies both at the observation and prediction levels. 

This work aims at generating a predictive model for video sequences. Our method can be inserted in a multi-modal architecture for autonomous vehicles. Future work is oriented to examine the incremental learning of the new scenarios and adaptation to them. Anomaly detection constitutes a fundamental aspect for this: when abnormal situations are detected, the corresponding input images can be used for building/refining new and already existing models. As discussed in the paper, two main abnormal situations can be distinguished: \textit{i)} Video sequences containing images that were never seen before; \textit{ii)} Video sequences containing known images but new video dynamics/motions. This paper mainly focused on the first case, as also done by \cite{Ravanbakhsh2018a, Ravanbakhsh2018b}. In the future, we will consider both cases by exploiting the incremental learning of new situations by taking abnormal data and use it for building/refining predictive models. 

The insertion of observations coming from other sensory data, e.g., positional and control information, into the proposed probabilistic framework constitutes another future path for our work. By allowing algorithms to handle multi-modal data when making predictions and detecting anomalies, it is possible to generate more realistic algorithms that associate multiple heterogeneous signals to familiar concepts as humans and other animals do. Such will certainly enable a more robust inference process, which leads to more efficient decision-making in autonomous systems.     

\bibliographystyle{IEEEtran}
\bibliography{IEEEfull}
\end{document}